\pdfoutput=1

\documentclass[11pt]{article}
\usepackage{float}
\usepackage[preprint]{acl}
\usepackage{booktabs,tabularx,array}
\newcolumntype{Y}{>{\raggedright\arraybackslash}X} 

\usepackage{times}
\usepackage{latexsym}
\usepackage{svg}

\usepackage{algorithmicx}
\usepackage{algcompatible}
\usepackage[T1]{fontenc}
\usepackage{algpseudocode}
\usepackage[utf8]{inputenc}
\usepackage{booktabs}
\usepackage{tabularx}
\usepackage{makecell} 
\usepackage{microtype}

\usepackage{graphicx}
\usepackage{amsmath}

\usepackage{amsmath}          
\usepackage{algorithm}        
\usepackage{booktabs}
\usepackage{array}
\usepackage{tabularx}
\usepackage{enumitem}
\usepackage{algorithm}
\usepackage{multirow}
\title{Patent Language Model Pretraining with ModernBERT}
\author{
 \textbf{Amirhossein Yousefiramandi} \hspace{0.2em}
 \textbf{Ciarán Cooney} \hspace{0.2em}
\\[1ex]
 Clarivate
\\
   \\[-1.0em]
\small{
   \textbf{Correspondence: }amirhossein.yousefiramandi@clarivate.com
 }
}

\begin{document}
\maketitle
\begin{abstract}
Transformer-based language models such as BERT have become foundational in NLP, yet their performance degrades in specialized domains like patents, which contain long, technical, and legally structured text. Prior approaches to patent NLP have primarily relied on fine-tuning general-purpose models or domain-adapted variants pretrained with limited data. In this work, we pretrain 3 domain-specific masked language models for patents, using the ModernBERT architecture and a curated corpus of over 60 million patent records. Our approach incorporates architectural optimizations, including FlashAttention, rotary embeddings, and GLU feed-forward layers. We evaluate our models on four downstream patent classification tasks. Our model, ModernBERT-base-PT, consistently outperforms the general-purpose ModernBERT baseline on three out of four datasets and achieves competitive performance with a baseline PatentBERT. Additional experiments with ModernBERT-base-VX and Mosaic-BERT-large demonstrate that scaling the model size and customizing the tokenizer further enhance performance on selected tasks. Notably, all ModernBERT variants retain substantially faster inference—over 3× that of PatentBERT—underscoring their suitability for time-sensitive applications. These results underscore the benefits of domain-specific pretraining and architectural improvements for patent-focused NLP tasks.
\end{abstract}

\section{Introduction}
Since its release in 2018, BERT and other derivative encoder-based transformer models have become a mainstay of modern NLP research and applications ~\cite{devlin2019bert}. Even within an ecosystem apparently dominated by generative Large Language Models (LLMs), BERT-based models continue to be used widely in both traditional encoder applications as well as within some LLM systems e.g., Retrieval Augmented Generation systems ~\cite{lewis2020retrieval, wang2022text}. 

Optimized primarily for general-domain text, BERT tends to underperform in specialized domains that feature distinct linguistic structures, such as legal or biomedical corpora \cite{beltagy2019scibert, limsopatham2021effectively}. Many such domains—including patents—exhibit idiosyncratic lexical and syntactic features, warranting domain-specific adaptation. While fine-tuning on downstream tasks remains a popular strategy, several studies have demonstrated that extending pretraining on domain-relevant corpora, or pretraining from scratch, can lead to substantial performance gains \cite{chalkidis2020legal, rasmy2021med}.

Patent documents are a unique blend of legal language and technical exposition, often structured and written in ways that differ sharply from general web or news text. Tasks in this domain—including classification, retrieval, and paragraph highlighting—have been addressed using both traditional ML \cite{kamateri2022automated, haghighian2022patentnet} and transformer-based approaches \cite{lee2020patent, bekamiri2024patentsberta}. However, the majority of BERT-based studies in this space rely on generic pretraining, which may limit their effectiveness given the domain shift.

To date, only one notable effort has reported on pretraining a BERT-style model specifically for patent documents \cite{srebrovic2020leveraging}. This indicates a significant gap in the literature, especially given the demonstrated benefits of domain adaptation in comparable fields \cite{chalkidis2020legal, qiu2021u, yan2022clinical}. Our work aims to address this gap directly. Additionally, several technical enhancements to BERT have recently been proposed and validated, including FlashAttention ~\cite{dao2022flashattention, shah2024flashattention}, unpadding ~\cite{zeng2022boosting}, optimized MLM masking ~\cite{wettig2023should}, and rotary positional embeddings ~\cite{su2024roformer}. These improvements not only accelerate training, but also improve model quality and stability ~\cite{portes2023mosaicbert, warner2024smarter}.

In this work, we build on these architectural and training innovations by using the ModernBERT design in conjunction with MosaicML’s Composer framework to pretrain a transformer model specifically for patent-domain tasks. Our contributions are as follows:
\begin{itemize}
    \item \textbf{Architecture and Efficiency:} We adopt the ModernBERT architecture and incorporate recent training optimizations, such as FlashAttention \cite{dao2022flashattention}, dynamic masking ratios \cite{wettig2023should}, and rotary positional embeddings \cite{su2024roformer}, as facilitated by MosaicML's Composer framework. These enhancements significantly reduce pretraining time and compute cost compared to traditional BERT, making the model both more accessible and scalable.
    \item \textbf{Training Data:} We construct a diverse and representative corpus that blends public patent data with proprietary patent text. The proprietary data are sections of original patent text that have been rewritten in less obscure language by subject-matter experts. This hybrid dataset enhances the model’s ability to generalize across patent categories and jurisdictions.
    \item \textbf{Downstream Evaluation:} We benchmark our pretrained model on multiple downstream patent classification tasks and compare it against both a generic ModernBERT base model and Google’s PatentBERT \cite{srebrovic2020leveraging}. Our results demonstrate that pretraining on domain-specific patent data improves task accuracy over ModernBERT-base and leads to more efficient fine-tuning and inference in comparison with PatentBERT. Moreover, results with ModernBERT-base-VX and Mosaic-BERT-large indicate that both architectural scaling and tokenizer customization can yield further gains on selected datasets, suggesting complementary strategies for optimizing patent-domain transformers.
\end{itemize}

\section{Methods}

\subsection{Data and Preprocessing}
\label{sec:arch}

Pretraining a domain-specific language model with MLM loss requires a substantial volume of domain-specific data in order to compete with or exceed the performance of more generic pretrained encoder models. We construct our dataset from \textasciitilde100M patent documents (mean length \textasciitilde700 words), comprising original published applications and corresponding proprietary samples extracted from the Derwent World Patents Index (DWPI)\footnote{\url{https://clarivate.com/intellectual-property/patent-intelligence/derwent-world-patents-index/}}. This DWPI data has been constructed by subject matter experts to provide more insightful descriptions of patent content.

To prepare the data, we applied several preprocessing and quality enhancement steps. From the broader corpus, we extracted abstracts and the first independent claim from each patent, and appended DWPI titles, abstracts, and claims by matching publication numbers. Non-essential text (e.g., figure references, non-English characters) was removed. Only the first independent claim per patent was retained.

Following strategies used in FineWeb ~\cite{penedo2024fineweb}, we applied a language filter to retain only English patents and implemented text quality and repetition filters ~\cite{rae2021scaling}, using the DataTrove toolkit\footnote{\url{https://github.com/huggingface/datatrove}}~\cite{penedo2024fineweb}. Additional deduplication was critical given that patents are often clustered in families or legal hierarchies: we applied MinHash-based fuzzy deduplication ~\cite{penedo2024fineweb}, which computes n-gram signatures and filters out duplicates or near-duplicates. This procedure reduced the dataset from \textasciitilde100M to \textasciitilde64M unique patents.

Our final training dataset contains \textasciitilde64M patents (\textasciitilde30.8B tokens), with a held-out test set of 3.4M patents (\textasciitilde1.6B tokens). This training corpus is approximately half the size of Google PatentBERT \cite{srebrovic2020leveraging}. Full preprocessing details are reported in Appendix ~\ref{app:preprocessing}.

\label{sec:transformerplusplus}
\subsection{Tokenization and Vocabulary Construction}
\label{sec:transformerplusplus}
To support MLM pre-training from scratch on English-language patent text, we developed a domain-specific subword tokenizer for MosaicBERT-large (~\ref{sec:efficiency}) using the tokenizers library \cite{wolf2020transformers}. The tokenizer implements a \emph{Byte-Pair Encoding} (BPE) model \cite{sennrich2016subword} initialized with an unknown token \texttt{[UNK]}. All input text is normalized by Unicode NFKC normalization followed by lowercasing, reducing case-induced sparsity and consolidating visually or canonically distinct code points. Pre-tokenization uses whitespace splitting; punctuation remains attached to neighboring tokens but can be segmented by the subword model as needed.

The BPE vocabulary was trained from a streaming iterator over the pretraining corpus with a target size of 49,152 subwords and a minimum frequency threshold of 2. The following special tokens were reserved for MLM training: \texttt{[UNK]}, \texttt{[PAD]}, \texttt{[CLS]}, \texttt{[SEP]}, and \texttt{[MASK]}. The final vocabulary and merge rules were serialized and reused for all pretraining and evaluation. At inference time, documents are normalized, whitespace-tokenized, and mapped to subword indices; frequent technical terms and morphemes appear as single tokens, while rarer terms decompose into multiple subwords. Out-of-vocabulary character sequences are mapped to \texttt{[UNK]}.

For comparison, we also experimented with WordPiece-style tokenization \cite{schuster2012wordpiece,devlin2019bert}, including section-aware variants with special markers for abstracts, claims, and DWPI text \cite{srebrovic2020bertpatents}. These tokenizers were trained with varying hyperparameters and evaluated using average token entropy \cite{zouhar2023tokenization,dagan2024tokenizer} and unknown-token rates. However, preliminary end-to-end experiments showed weaker downstream performance relative to our BPE tokenizer. Since detailed results are out of scope, and all models reported in this paper use either our BPE tokenizer or ModernBERT defaults, we do not report further on these WordPiece experiments.

The other two pretrained models, ModernBERT-base-PT and ModernBERT-base-VX, employ their original publicly released tokenizers \cite{warner2024smarter} without modification.

\paragraph{Model suffixes (corpus variants).}
We suffix model names to indicate the pretraining corpus variant (Table \ref{tab:model-suffix-legend}). 
\textbf{\texttt{-PT}} (\emph{clean}) denotes pretraining on the Phase~2 cleaned and near-deduplicated patent corpus, which applies English language identification, repetition/quality filters, a FineWeb-style quality filter, and MinHash-based near-duplicate removal (Appendix \ref{app:preprocessing:B2}). 
\textbf{\texttt{-VX}} (\emph{raw}) denotes pretraining on the corpus prior to Phase~2 filtering—i.e., after Phase~1 extraction/normalization and family-level deduplication only (Appendix \ref{app:preprocessing:B1}). 
The raw variant is larger but noisier: Phase~2 reduces tokens from \(\sim 47.7\)B tokens for \textit{ModernBERT-base-VX} to \(\sim 31.6\)B tokens for \textit{ModernBERT-base-PT} (\(-33.79\%\); Appendix \ref{tab:reductionss}). See Appendix \ref{app:training}.

\begin{table*}[t]
\centering
\footnotesize
\setlength{\tabcolsep}{6pt}
\begin{tabularx}{\textwidth}{@{}Y c Y c X@{}}
\toprule
\textbf{Model} & \textbf{Sfx} & \textbf{Corpus} & \textbf{Tokens} & \textbf{Definition / Notes} \\
\midrule
ModernBERT-base-PT & \texttt{-PT} & Clean (Phase~2) & $\sim$31.6B &
Phase~2 cleaned \& near-deduplicated corpus: English LID; MassiveText/Gopher repetition/quality filters; FineWeb; MinHash (Appendix \ref{app:preprocessing:B2}). Phase‑2 total $\approx$31.64B (Appendix \ref{tab:reductionss}). \\
ModernBERT-base-VX & \texttt{-VX} & Raw (pre–Phase~2) & $\sim$47.7B &
Pre‑Phase~2 corpus (after Phase~1 \& family‑level dedup only; Appendix \ref{app:preprocessing:B1}). More tokens, higher noise; aligns with $\approx$47.79B (Appendix \ref{tab:reductionss}). \\
MosaicBERT-large (BPE) & — & Raw (pre–Phase~2)\textsuperscript{$\dagger$} & $\sim$47.7B &
Custom BPE tokenizer (§\ref{sec:transformerplusplus}); long context; trained on $\sim$47.7B tokens (Appendix \ref{app:training}). \\
\bottomrule
\end{tabularx}
\caption{\textbf{Model suffix legend and corpus variants.} \texttt{-PT} uses the Phase~2 \emph{clean} corpus; \texttt{-VX} uses the \emph{raw} corpus prior to Phase~2.}
\label{tab:model-suffix-legend}
\end{table*}

\subsubsection{Model Architecture}
\label{sec:efficiency}
We adopt the FlexBERT-base architecture from the \textit{ModernBERT} repository\footnote{\url{https://github.com/AnswerDotAI/ModernBERT}}, which is a modernized BERT-based encoder tailored for efficient training \cite{warner2024smarter}. This architecture consists of 22 encoder blocks with a hidden state size of 768 and 12 self-attention heads per layer. Gaussian Error Linear Unit (GELU) is the activation function for all feed-forward layers and Layer Normalization is applied at the beginning of each sublayer to improve training stability. The original ModernBERT has a maximum sequence length of 8192 tokens but we have limited this to 1024 for our requirements.  

Following MosaicBERT \cite{portes2023mosaicbert}, our implementation integrates several efficiency optimizations. First, the model uses FlashAttention for the self-attention layers, enabling larger batch sizes and longer sequences \cite{dao2022flashattention, shah2024flashattention}. Second, we adopt ALiBi (Attention with Linear Biases) positional encoding instead of learned positional embeddings \cite{press2021train}. ALiBi adds a relative position bias to attention scores rather than using absolute position embeddings, which not only improves final model performance but also enables the model to generalize to sequences longer than those seen in training. Third, feed-forward layers are implemented with a Gated Linear Unit (GLU)  \cite{portes2023mosaicbert}. This implementation sees the intermediate dense layer’s output split into two parts. One part is passed through a GELU activation and the other part is used as a gating signal (after a linear transformation) that multiplicatively modulates the activated part, enabling more flexible and expressive representations. Fourth, we scale residual connections by a factor of 0.5 following the Pre-LayerNorm setup, which stabilizes training in deeper architectures. Notably, dropout is removed entirely from the architecture without adverse effects on generalization or training stability. These model enhancements have been shown to improve training speed and stability without sacrificing model performance \cite{portes2023mosaicbert}.

In addition to this baseline, we pretrained two further model variants.
First, a ModernBERT-base-VX configuration, which retains 22 layers and 12 attention heads but introduces adjusted intermediate dimensions and GLU expansion consistent with the Composer framework. Specifically, this variant uses a hidden size of 768, intermediate size of 1152, and GLU expansion to 2304, with RoPE $\theta$ set to 160,000 and a local attention RoPE $\theta$ of 10,000. While the total parameter count remains approximately 149M, training differed slightly from the baseline in terms of optimizer configuration and batch scheduling (see Appendix ~\ref{app:training}). This provides a controlled point of comparison against our domain-pretrained ModernBERT-base-PT.

Second, a Mosaic-BERT-large model was trained using our custom BPE tokenizer (section~\ref{sec:efficiency}). This model scales to 28 encoder layers with a hidden size of 1024 and 16 attention heads, yielding a total of approximately 395M parameters. The feed-forward intermediate dimension is expanded to 2624, with GLU expansion to 5248. Like the base variants, it integrates FlashAttention, rotary embeddings, and Pre-LayerNorm residual scaling, but supports a maximum sequence length of 8192 tokens, making it particularly suited for patent texts with long contexts. Training followed a trapezoidal learning rate schedule with StableAdamW, distributed across 8×H100 GPUs, with dropout applied only to attention outputs (0.1). This model allows us to assess both the effect of scaling and the contribution of tokenizer customization.

\begin{figure*}[t]
    \centering
    \includegraphics[width=\textwidth]{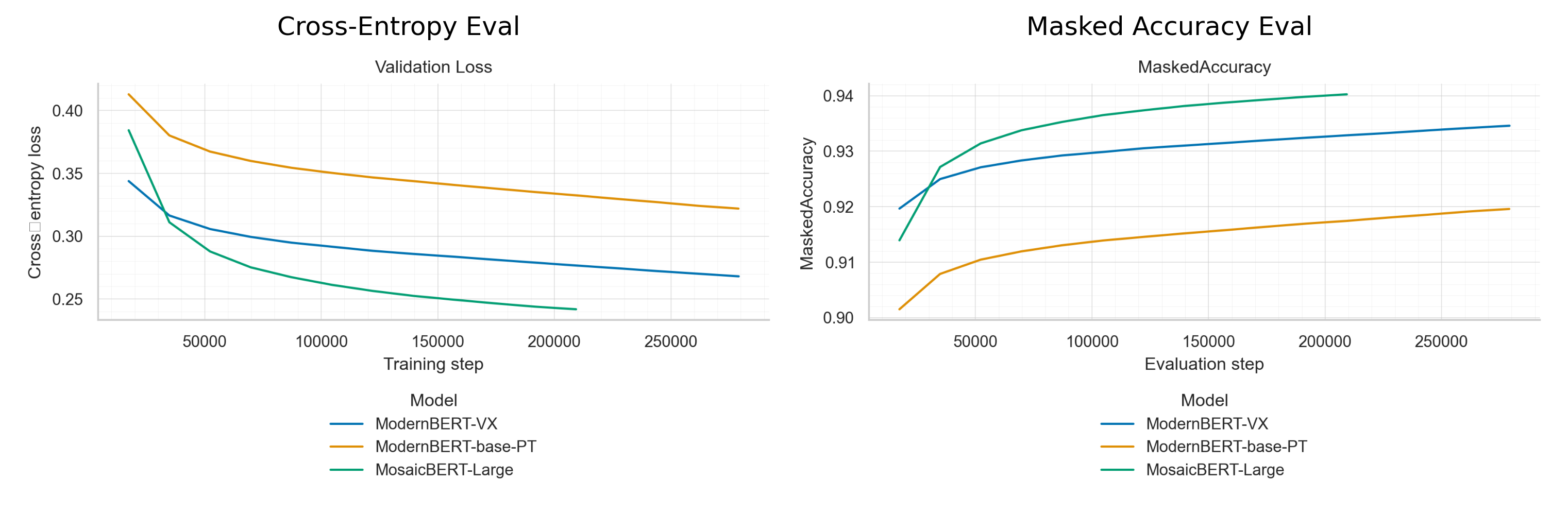}
    \caption{Pretraining loss and MLM accuracy for ModernBERT-base-PT, ModernBERT-base-VX and Mosaic-BERT-large.}
    \label{fig:pretrain_loss}
\end{figure*}

\subsubsection{Training Procedure}
\label{sec:efficiency}
Pretraining used the MLM objective. We did not include the Next Sentence Prediction task as it has been shown to provide minimal downstream performance improvement \cite{liu2019roberta, izsak2021train}. We used a masking probability of 30\% as recent findings indicate that the original probability of 15\% used in BERT is suboptimal \cite{wettig2023should}. Training was optimized using StableAdamW \cite{wortsman2023stable}, an enhancement of AdamW \cite{loshchilovdecoupled} that incorporates update clipping on a per-parameter basis similar to Adafactor \cite{shazeer2018adafactor}. Hyperparameter details are provided in Appendix ~\ref{app:hyperparameters}.

To avoid initial instability the learning rate schedule employed used a linear warmup followed by linear decay. Specifically, the learning rate was linearly ramped up from 0 to $3 \times 10^{-4}$ over the first 6\% of training steps during a warmup period and then linearly decayed for the remaining 94\% of training steps, reaching about 2\% of the peak learning rate by the end of training. We use a global batch size of 4096 sequences per step during training. This required each GPU processing a micro-batch of 128 sequences, with gradients synchronized across GPUs prior to each optimizer update. The maximum sequence length during training was set to 1024 tokens per example.



\section{Results}

\begin{figure*}[t]
    \centering
    \includegraphics[width=\textwidth]{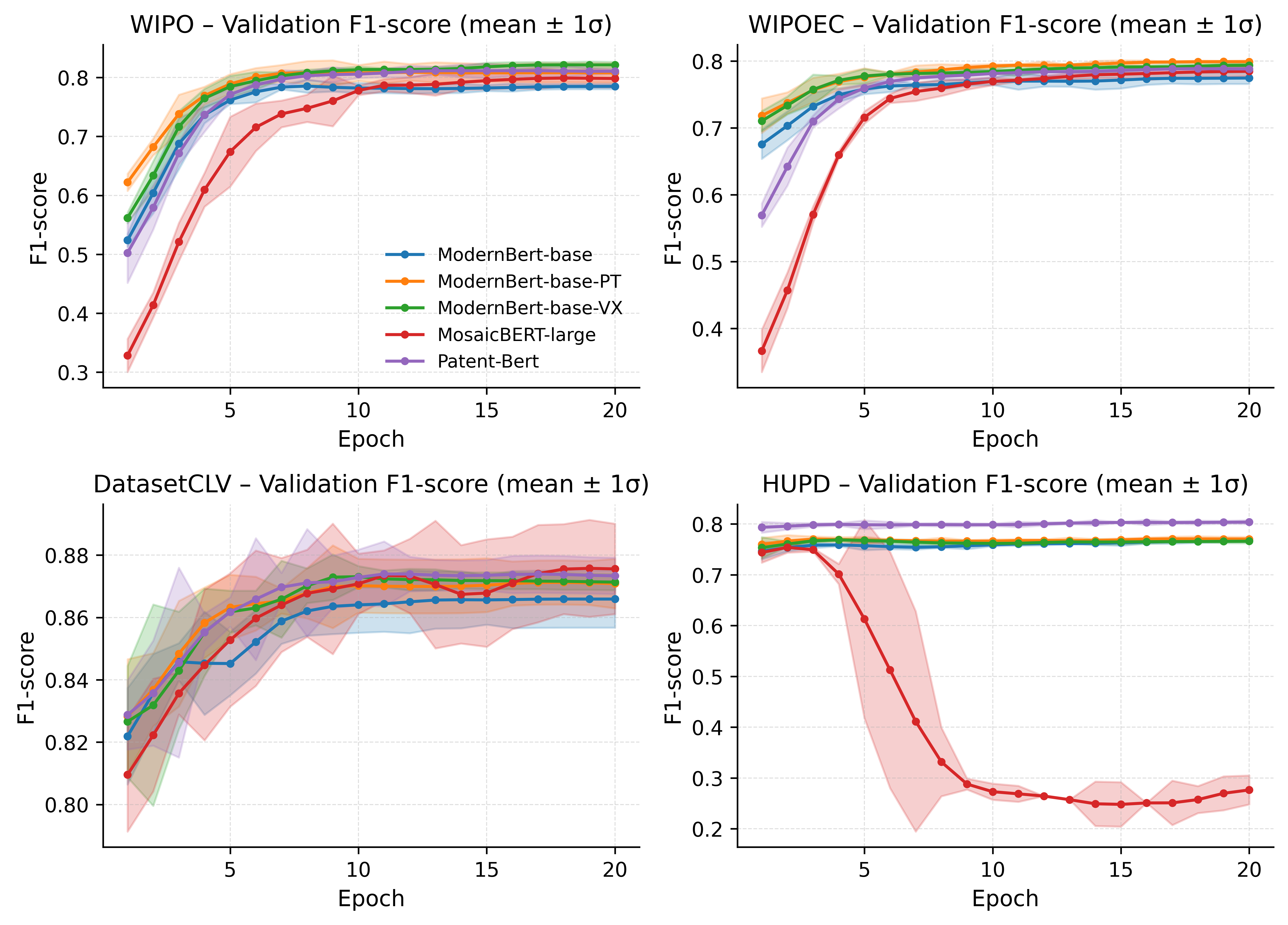}
    \caption{Validation f1-scores for finetuning tasks across 20 epochs.}
    \label{fig:example-wide}
\end{figure*}

\begin{table}[t]
\centering
\setlength{\tabcolsep}{5pt} 
\begin{tabular}{lrrrr}
\toprule
\textbf{Dataset} & \textbf{Labels} & \textbf{Train} & \textbf{Val.} & \textbf{Test} \\
\midrule
WIPO       & 14 & 1731  & 424  & 533  \\
WIPOEC     & 43 & 11349 & 2798 & 3524 \\
HUPD       &  8 & 20772 & 5194 & 6000 \\
DatasetCLV &  5 & 1481  & 371  & 400  \\
\bottomrule
\end{tabular}
\caption{Dataset statistics for downstream evaluation tasks.}
\label{tab:datasets}
\end{table}

\subsection{Pretraining Evaluation}
Figure~\ref{fig:pretrain_loss} presents the training loss and masked accuracy for ModernBERT-base-VX, ModerBERT-base-PT and MosaicBERT-large tokenizer. The model trained with the custom BPE tokenizer exhibited a steeper decline in MLM loss in the early stages of pretraining and maintained a consistently lower loss throughout training (Figure~\ref{fig:pretrain_loss}). Similarly, masked token prediction accuracy rose more sharply and stabilized at a higher level compared to the WordPiece-based counterparts (Figure~\ref{fig:pretrain_loss}).

Pretraining results suggest that the BPE tokenizer provided a more efficient and compact representation of the patent text. BPE is known for its ability to balance vocabulary size and token granularity, particularly by effectively encoding frequent subword patterns and maintaining coherence for rare or domain-specific terms. In contrast, the standard WordPiece tokenizer—optimized for general-domain corpora—often splits complex technical terms into less informative units, making masked token prediction more difficult.




\subsection{Downstream Evaluation}
\begin{figure*}[t]
    \centering
    \includegraphics[width=\textwidth]{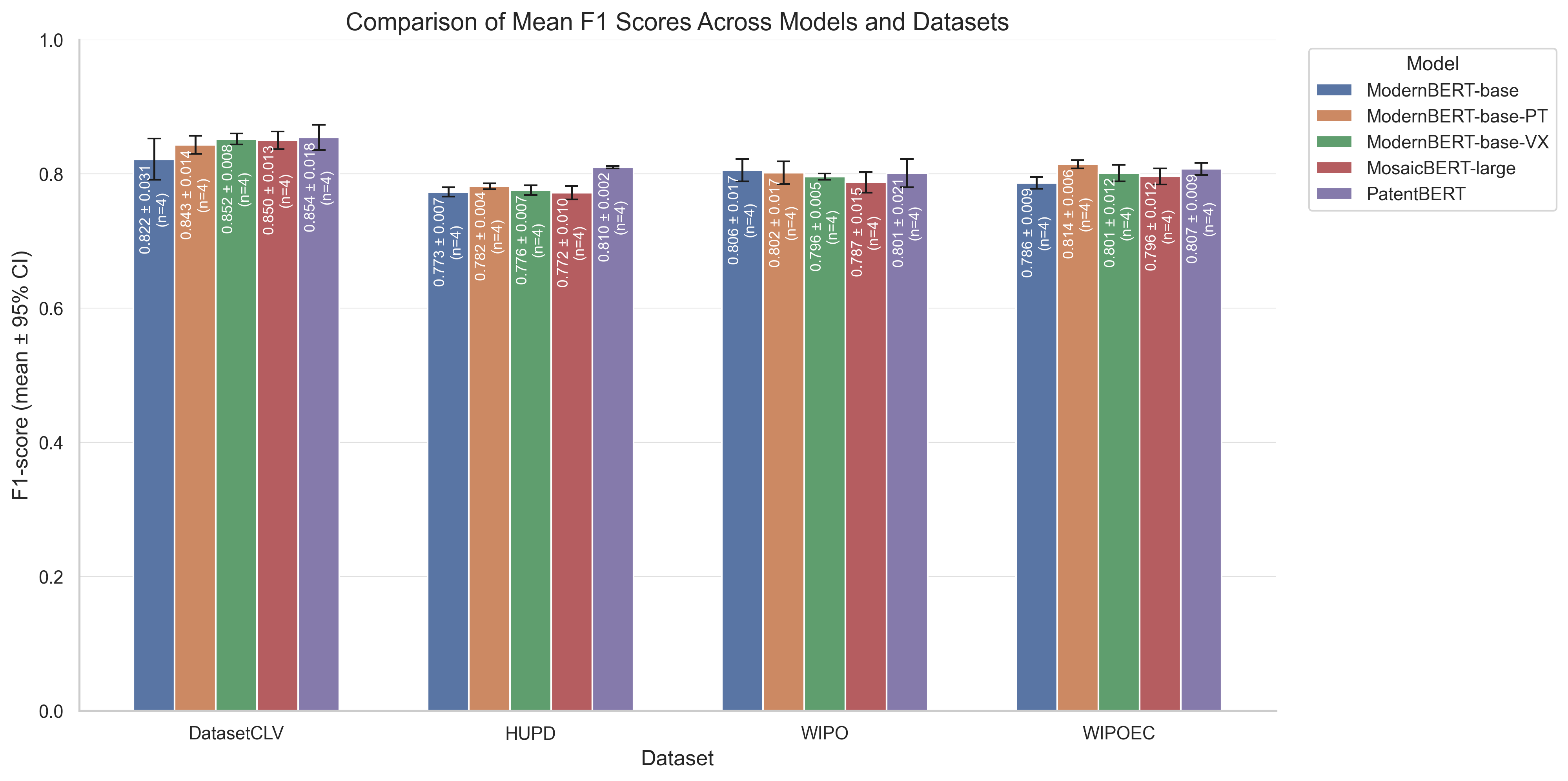}
    \caption[Test micro-F1 across models and datasets]{\textbf{Test} \textbf{micro-averaged F1-scores} across models and datasets, reported as mean \(\pm\) 95\% confidence interval computed over \textbf{4 random seeds} per (dataset, model). Confidence intervals are estimated using the \textit{Student's \(t\)} distribution and clipped to \([0,1]\) to respect the bounded range of F1.}
    \label{fig:all-datasets-models-f1}
\end{figure*}

We validated our pretrained patent language models on downstream classification tasks using four different datasets (Table~\ref{tab:datasets}; Figure~\ref{fig:all-datasets-models-f1} with confidence intervals). The first two datasets come from the World International Patent Office (WIPO)\footnote{\url{https://www.wipo.int/portal/en/index.html}}. Another dataset is the Harvard USPTO Patent Dataset (HUPD) \cite{suzgun2023harvard}\footnote{\url{https://huggingface.co/datasets/HUPD/hupd}}. The final dataset is a proprietary dataset consisting of 5 category labels relating to data storage and networking technology. Each was split into train/validation/test sets with experiments run four times using different seeding. Reported f1 scores in  Table~\ref{tab:model-results} are averages of these experiments. Details on the category labels of each of the three open datasets can be found in Appendix ~\ref{app:datainfo}.

ModernBERT-base-PT improved upon the performance of ModernBERT-base on three of the four datasets. ModernBERT-base-VX achieved broadly similar results, with slightly higher performance on the DatasetCLV dataset. Mosaic-BERT-large, trained with our custom BPE tokenizer, also demonstrated strong performance, though without consistently surpassing the smaller domain-pretrained models. PatentBERT remained strongest on HUPD, reflecting its use of USPTO data during pretraining. Taken together, these results indicate that domain-specific pretraining, architectural scaling, and tokenizer customization each provide complementary advantages for patent-specific NLP, though no single strategy yet establishes a decisive state of the art across all benchmarks.


In figure ~\ref{fig:example-wide}, evaluation f1 scores across 20 epochs of finetuing are shown (evaluation loss plots are in Appendix ~\ref{app:fullresults}). All four plots exhibit our ModernBERT-base-PT with higher evaluation f1 scores in comparison with ModernBERT-base at all data points. This suggests there is value in domain-specific pretraininig for patent-specific downstream tasks. The plot depicting the HUPD f1 scores is noteworthy as PatentBERT outperforms the other models by a degree not seen in the other plots. We hypothize the reason for this is that USPTO data has been used during the pretraining of PatentBERT.

\begin{table*}[t]
\centering
\begin{tabular}{lcccc}
\toprule
\textbf{Model} & \textbf{WIPO} & \textbf{WIPOEC} & \textbf{HUPD} & \textbf{DatasetCLV}\\
\midrule
ModernBERT-base & \textbf{0.806} $\pm$ 0.011 & 0.786 $\pm$ 0.005 & 0.773 $\pm$ 0.004 & 0.822 $\pm$ 0.019 \\
ModernBERT-base-PT & 0.802 $\pm$ 0.011 & \textbf{0.814} $\pm$ 0.004 & 0.782 $\pm$ 0.003 & 0.843 $\pm$ 0.009 \\
ModernBERT-base-VX & 0.796 $\pm$ 0.003 & 0.801 $\pm$ 0.008 & 0.776 $\pm$ 0.005 & 0.852 $\pm$ 0.005 \\
MosaicBERT-large & 0.787 $\pm$ 0.010 & 0.796 $\pm$ 0.007 & 0.772 $\pm$ 0.006 & 0.850 $\pm$ 0.008 \\
PatentBERT & 0.801 $\pm$ 0.013 & 0.807 $\pm$ 0.006 & \textbf{0.810} $\pm$ 0.001 & \textbf{0.854} $\pm$ 0.012 \\
\bottomrule
\end{tabular}
\caption{F1-score (mean $\pm$ std across 4 runs) for downstream classification tasks on four patent datasets.}
\label{tab:model-results}
\end{table*}

\subsection{Finetuning Efficiency}
Consistent with prior reports of ModernBERT’s efficiency~\cite{warner2024smarter}, our measurements corroborate high throughput across four patent datasets (WIPO, WIPOEC, HUPD, DatasetCLV). Per-dataset mean~$\pm$~std over four seeds is reported for inference (Table~\ref{tab:model-infer-throughput}) and training (Table~\ref{tab:model-train-throughput}); confidence-interval plots appear in Appendix~\ref{app:fullresults}. \textit{Implementation details.} In our setup, \textbf{ModernBERT-base}, \textbf{ModernBERT-base-PT}, and \textbf{ModernBERT-base-VX} use \emph{FlashAttention}; \textbf{MosaicBERT-large} uses \emph{ALiBi} positional bias with non-FlashAttention kernels; and \textbf{PatentBERT} uses a pre-FlashAttention (vanilla scaled dot-product) attention stack. \textit{Aggregate speed.} Macro-averaged across datasets, the FlashAttention-based ModernBERT family attains \textbf{2.32$\times$} higher training throughput and \textbf{3.55$\times$} higher inference throughput than PatentBERT (\,+132\% and +255\%, respectively), and \textbf{2.89$\times$} / \textbf{2.93$\times$} over MosaicBERT-large in training/inference (\,+189\% / +193\%). Looking at individual variants, \textbf{ModernBERT-base} is the fastest in training, running \textbf{1.48$\times$} and \textbf{1.40$\times$} faster than ModernBERT-base-PT and ModernBERT-base-VX, respectively; inference among the three ModernBERT models is essentially tied (PT and VX are $\approx$\textbf{1.02$\times$} ModernBERT-base). Relative to non-Flash baselines, ModernBERT-base is \textbf{3.63$\times$} faster in training and \textbf{2.90$\times$} in inference than MosaicBERT-large, and \textbf{2.91$\times$} / \textbf{3.51$\times$} faster than PatentBERT (training/inference). These results support FlashAttention-based ModernBERT as a competitive alternative to other BERT variants, particularly for time-sensitive inference. \emph{Note:} Throughput differences reflect both attention kernels and parameter counts (149M for ModernBERT vs.\ 340–346M for MosaicBERT/PatentBERT), so part of the gain is architectural while part is model size.



\begin{table*}[t]
\centering
\begin{tabular}{lccccc}
\toprule
\textbf{Model} & \textbf{Parameters} & \textbf{WIPO} & \textbf{WIPOEC} & \textbf{HUPD} & \textbf{DatasetCLV}\\
\midrule
ModernBERT-base-PT & 149M & 83.68 $\pm$ 0.09 & \textbf{85.54 $\pm$ 0.15} & 368.42 $\pm$ 0.33 & 90.99 $\pm$ 0.83 \\
PatentBERT & 346M & 36.09 $\pm$ 0.06 & 35.50 $\pm$ 0.12 & 68.84 $\pm$ 0.38 & 35.51 $\pm$ 0.05 \\
ModernBERT-base & 149M & \textbf{84.48 $\pm$ 0.04} & 62.66 $\pm$ 0.52 & \textbf{377.41 $\pm$ 1.73} & \textbf{93.45 $\pm$ 0.42} \\
ModernBERT-base-VX & 149M & 84.06 $\pm$ 1.04 & 84.51 $\pm$ 0.38 & 368.09 $\pm$ 0.75 & 91.87 $\pm$ 0.23 \\
MosaicBERT-large & 340M & 27.44 $\pm$ 0.09 & 26.75 $\pm$ 0.56 & 127.39 $\pm$ 0.60 & 31.74 $\pm$ 0.30 \\
\bottomrule
\end{tabular}
\caption{Inference throughput (samples/sec; mean $\pm$ std over runs) for four patent datasets.}
\label{tab:model-infer-throughput}
\end{table*}

\begin{table*}[t]
\centering
\begin{tabular}{lccccc}
\toprule
\textbf{Model} & \textbf{Parameters} & \textbf{WIPO} & \textbf{WIPOEC} & \textbf{HUPD} & \textbf{DatasetCLV}\\
\midrule
ModernBERT-base-PT & 149M & 21.11 $\pm$ 0.43 & 21.48 $\pm$ 1.01 & 24.98 $\pm$ 1.58 & 23.33 $\pm$ 0.16 \\
PatentBERT & 346M & 8.65 $\pm$ 0.14 & 9.29 $\pm$ 0.06 & 19.29 $\pm$ 0.56 & 8.87 $\pm$ 0.07 \\
ModernBERT-base & 149M & \textbf{21.51 $\pm$ 0.54} & \textbf{23.17 $\pm$ 0.93} & \textbf{65.04 $\pm$ 2.13} & \textbf{24.55 $\pm$ 0.20} \\
ModernBERT-base-VX & 149M & 20.85 $\pm$ 0.38 & 21.15 $\pm$ 0.80 & 30.38 $\pm$ 2.67 & 23.50 $\pm$ 0.13 \\
MosaicBERT-large & 340M & 6.19 $\pm$ 0.12 & 6.53 $\pm$ 0.16 & 17.33 $\pm$ 0.71 & 6.92 $\pm$ 0.07 \\
\bottomrule
\end{tabular}
\caption{Training throughput (samples/sec; mean $\pm$ std over runs) for four patent datasets.}
\label{tab:model-train-throughput}
\end{table*}

\section{Limitations}

\textbf{Language} 
Our pretraining corpus is limited to English-language patents. Given the global nature of patent filings, expanding to multilingual corpora could significantly improve model utility in international contexts.

\textbf{MLM-only objective}
Following \citeauthor{warner2024smarter}, we rely solely on the MLM objective for pretraining. While MLM has proven effective, it does not capture sentence-level semantics or inter-document relationships. Future work could explore complementary pretraining tasks such as contrastive objectives, span prediction, or retrieval-based learning tailored for patent data.

\textbf{Scaling}
Our model was trained on a compute budget significantly smaller than that of large-scale commercial or academic models. While this demonstrates efficiency, it may limit representational capacity compared to larger-scale alternatives. Additionally, despite pretraining on ~60M patents, there are potential performance increases available from further scaling of the training data.

\textbf{Limited Downstream Tasks}
Our downstream evaluation focuses on patent classification tasks. While results are strong, they do not yet demonstrate generalization to other patent-related tasks such as similarity ranking, summarization, claim matching, or novelty detection.

\section{Conclusion}
In this work, we presented ModernBERT-PT, a domain-specific masked language model pretrained from scratch on a curated corpus of over 60 million patent documents. Leveraging architectural innovations such as FlashAttention, ALiBi positional embeddings, and GLU-based feed-forward layers, we demonstrated that pretraining on patent-specific data yields tangible benefits in both classification accuracy and computational efficiency. ModernBERT-base-PT outperformed a general-purpose ModernBERT baseline on three out of four downstream classification tasks and achieved competitive performance with PatentBERT, while offering substantially faster inference speeds. Additional comparisons with ModernBERT-base-VX and Mosaic-BERT-large suggest that scaling model capacity and introducing customized tokenization provide further gains on certain datasets, particularly those with longer or more technical contexts. These findings highlight the value of combining domain-specific pretraining with targeted architectural and tokenization strategies for advancing patent-domain NLP.

These findings highlight the importance of domain-specific pretraining and architecture optimization for specialized text corpora like patents. While our results validate ModernBERT-base-PT as a strong foundation for patent NLP tasks, several avenues remain for future improvement. These include scaling up the training corpus, incorporating multilingual patents, and expanding evaluations to a broader set of downstream tasks such as retrieval, summarization, and novelty detection.

With repsect to tokenization, results support growing evidence that tokenizer customization, particularly with BPE, can significantly enhance pretraining efficiency and downstream performance in domain-specific NLP. Future work could explore comparisons between BPE, SentencePiece unigram models, or byte-level tokenization schemes to further isolate the effects of subword modeling strategies in specialized corpora like patents.
Overall, our work underscores the value of tailored model development for unlocking the full potential of NLP in technical and legally structured domains.

\section{Acknowledgements}
We would like to thank Clarivate for supporting this project and facilitating the compute resources. We would also like to thank Saurabh Mishra and Peter Keyngnaert for their support and feedback throughout the project.

\bibliography{custom}
\bibliographystyle{acl_natbib}

\appendix
\onecolumn

\section{Dataset Information}
\label{app:datainfo}
WIPO Vision Dataset (WIPO) \\
\textbf{Categories:}
\begin{itemize}[noitemsep, topsep=0pt]
\item Adaptive Focus
\item Artificial Iris
\item Artificial Silicon Retina (ASR) / Retinal Prostheses
\item Augmented Reality Devices
\item Bionic Eye (System)
\item Cortical Implants
\item Drug Delivery (Vision-related)
\item Hand Wearables
\item Intraocular Lenses (IOL) with Sensors
\item Intracorneal Lenses
\item Multifocal
\item Smart Eyewear
\item Telescopic Lenses
\item Virtual Reality Devices\\
\end{itemize}

\begin{flushleft}
WIPOEC Dataset\\
\end{flushleft}
\textbf{Categories:}
\begin{itemize}[noitemsep, topsep=0pt]
\item Accessories for changing body position or lifting persons – mounted in combination with a bathtub
\item Accessories for changing body position or lifting persons – mounted in combination with a toilet
\item Assistive Products for Animal Care
\item Assistive Products for Camping
\item Assistive Products for Creating Arts and Crafts
\item Assistive Products for Hunting and Fishing
\item Assistive Products for Play  
\item Assistive Products for Playing and Composing Music 
\item Assistive Products for Vertical Accessibility
\item Bathroom Accessories
\item Bathroom/Toilet Unit
\item Bathtub and Accessories
\item Beds and Their Accessories
\item Building Structural Components
\item Entry/Exit and Openings
\item Environment Alarms
\item Fall Detectors
\item Food Preparation
\item Golf
\item Handrails and Grab Bars
\item Laundry
\item Light Fixtures
\item Mountaineering
\item Other Furniture Accessories
\item Personal Emergency Alarm Systems and Medical Alert ID
\item Portable Travel Aids
\item Shower Unit and Accessories
\item Shower/Bathroom and Toilet Chairs
\item Sitting Arrangements and Their Accessories
\item Sports Wheelchairs
\item Storage
\item Swimming
\item Tables and Their Accessories
\item Tennis or Table Tennis
\item Toilet Seat and Accessories
\item Urinals
\item Wandering and Locating of Persons/Items
\item Wash Basin and Accessories
\item Winter Sports
\item Workplace and Domestic Safety
\item Workplace/Domestic Machinery
\item Workplace/Domestic Object Conveyance, Hoisting or Repositioning, Crane
\item Workplace/Domestic Object Securing, Gripping, Holding, Carrying and Handling\\
\end{itemize}
HUPD Dataset\\
\textbf{Categories:}
\begin{itemize}[noitemsep, topsep=0pt]
\item C: Chemistry; Metallurgy
\item A: Human Necessities
\item H: Electricity
\item G: Physics
\item B: Performing Operations; Transporting
\item E: Fixed Constructions
\item F: Mechanical Engineering; Lighting; Heating; Weapons; Blasting
\item D: Textiles; Paper\\
\end{itemize}
\begin{flushleft}
DatasetCLV\\
\end{flushleft}
\textbf{Categories:} Five categories related to data storage and networking.





\section{Data Preprocessing}
\label{app:preprocessing}

This appendix documents our two-phase preprocessing pipeline for large-scale patent texts used to pretrain masked language models. Phase~1 focuses on extraction, basic normalization, and family-level deduplication; Phase~2 performs FineWeb-inspired filtering and MinHash-based near-duplicate removal implemented with HuggingFace \emph{DataTrove} on Databricks.

\paragraph{Data fields.} For each record we retain: publication number, the \textit{abstract}, the \textit{first independent claim} (selected from the full claim set), and the \textit{DWPI}. Phase~2 is applied \textbf{per field} (abstract, first claim, DWPI) and results are re-assembled at the record level.


\subsection{Phase 1: Extraction, Normalization, and Family-Level Deduplication}
\label{app:preprocessing:B1}
\noindent\textbf{Scope.} We extract fields, remove markup (HTML/XML), normalize or drop corrupted characters, strip repetitive boilerplate phrases in abstracts/claims, and remove references to figures/images. We then deduplicate by patent \emph{family}: for each family we keep the most recent publication.

\noindent\textbf{Scale impact (documents).} From \textbf{$\approx$150\,M} English-language patents to \textbf{$\approx$70\,M} unique family representatives.

\begin{algorithm}[H]
\caption{Phase~1: Extract, Clean, and Family-Deduplicate (pseudocode)}
\label{alg:phase1}
\begin{algorithmic}[1]
\State \textbf{Input:} Patent DB with fields \(\{ \texttt{pub\_no}, \texttt{abstract}, \texttt{claims}, \texttt{DWPI}, \texttt{family\_id}, \texttt{pub\_date} \}\)
\State \textbf{Output:} Clean, family-deduplicated records \(\{ \texttt{pub\_no}, \texttt{abstract}^*, \texttt{first\_ind\_claim}^*, \texttt{DWPI}^* \}\)
\For{each record \(r\)}
  \State Extract \(\texttt{pub\_no}\), \(\texttt{DWPI}\), \(\texttt{abstract}\), and select \(\texttt{first\_ind\_claim}\) from \(\texttt{claims}\)
  \State Clean text fields: remove HTML/XML tags; normalize/drop corrupted characters
  \State Remove repetitive boilerplate prefixes; drop image/figure references
\EndFor
\State Group records by \(\texttt{family\_id}\); within each group, keep the record with the most recent \(\texttt{pub\_date}\)
\State \Return unified table with cleaned \(\texttt{abstract}^*, \texttt{first\_ind\_claim}^*, \texttt{DWPI}^*\)
\end{algorithmic}
\end{algorithm}

\subsection{Phase 2: FineWeb-Inspired Filtering and MinHash Near-Deduplication (Per Field)}
\label{app:preprocessing:B2}
\noindent\textbf{Filtering (Phase~2a).} For each field independently, we apply:
\begin{itemize}
    \item \textbf{Language filtering} via a fastText-based LID model (retain English only) \cite{joulin2016bag}.
    \item \textbf{Quality and repetition filters} inspired by MassiveText/Gopher heuristics \cite{rae2021scaling}.
    \item \textbf{FineWeb-style quality filter} to remove subtle low-quality artifacts \cite{penedo2024fineweb}.
\end{itemize}

\noindent\textbf{Near-duplicate removal (Phase~2b).} We use MinHash LSH over 5-gram shingles with 64-bit hashes and a 14$\times$8 banding scheme (14 buckets, 8 hashes per bucket) to cluster and remove near-duplicates, keeping one representative per cluster, following DataTrove components for signatures, bucketing, clustering, and filtering \cite{penedo2024datatrove}.

\noindent\textbf{Execution environment.} The pipeline runs on \textbf{Databricks} with parallel tasks and Parquet I/O to object storage for intermediate artifacts and logs.

\begin{algorithm}[H]
\caption{Phase~2: Filter and Near-Deduplicate Per Field (pseudocode)}
\label{alg:phase2}
\begin{algorithmic}[1]
\State \textbf{Input:} Phase~1 output (cleaned \texttt{abstract}, \texttt{first\_ind\_claim}, \texttt{DWPI})
\State \textbf{Output:} High-quality, near-deduplicated fields

\For{each field $f \in \{\text{\texttt{abstract}}, \text{\texttt{first\_ind\_claim}}, \text{\texttt{DWPI}}\}$}
  \For{each document $d$ in $f$}
    \If{$\operatorname{LanguageID}(d) \neq \text{English}$}
      \State discard
    \EndIf
    \If{$\operatorname{RepetitionHeuristics}(d) = \text{fail}$}
      \State discard
    \EndIf
    \If{$\operatorname{QualityHeuristics}(d) = \text{fail}$}
      \State discard
    \EndIf
    \If{$\operatorname{FineWebStyleHeuristics}(d) = \text{fail}$}
      \State discard
    \EndIf
  \EndFor
  \State Compute MinHash signatures on 5-gram shingles (64-bit); LSH-bucket into $14 \times 8$ bands
  \State Cluster colliding documents; keep one representative per cluster; mark others as near-duplicates
\EndFor
\State Re-assemble fields at the record level; count tokens before/after deduplication
\end{algorithmic}
\end{algorithm}

\subsection{Summary of Reductions}
\label{tab:reductionss}
\begin{table}[H]
\centering
\caption{Dataset size reductions across phases.}
\label{tab:reductions}
\begin{tabular}{@{}lrrr@{}}
\toprule
\textbf{Phase} & \textbf{Before} & \textbf{After} & \textbf{Reduction} \\
\midrule
Phase~1 (patents) & $\approx$150{,}000{,}000 & $\approx$70{,}000{,}000 & $\approx$80{,}000{,}000 \\
Phase~2 (tokens)  & 47{,}791{,}818{,}029 & 31{,}644{,}981{,}330 & 16{,}146{,}836{,}699 \,\,(\textbf{--33.79\%}) \\
\bottomrule
\end{tabular}
\end{table}

\subsection{References for Filtering and Deduplication}

We follow the large-scale text cleaning and deduplication practices documented in FineWeb \cite{penedo2024fineweb}, MassiveText/Gopher \cite{rae2021scaling}, and use fastText language identification \cite{joulin2016bag}. Our implementation is built on the \emph{DataTrove} library \cite{penedo2024datatrove}.








\section{Training Recipe}
\label{app:training}

\begin{table}[h]
\small
\setlength{\tabcolsep}{4pt}      
\renewcommand{\arraystretch}{1.15} 
\centering
\begin{tabular}{lccccc}
\toprule
\textbf{Model} & \textbf{N tokens} & \textbf{Epochs} & \textbf{Train Time } & \textbf{Training Hardware } & \textbf{Training Strategy }\\
\midrule
ModernBERT-base-PT    &31.6B  &16  &2.352 day&8x H100 &Distributed DataParallel\\
ModernBERT-base-VX    &47.7B  &16  &3.264 day&8x H100&Distributed DataParallel \\
MosaicBERT-large      &47.7B  &12 &8.707 day&8x H100&Distributed DataParallel \\
\bottomrule
\end{tabular}
\caption{Training Statistics.}
\label{tab:hp-finetuning}
\end{table}

\section{Hyperparameters}
\label{app:hyperparameters}

\textbf{Pretraining Hyperparameters}\\


\begin{table}[H]
\centering
\small
\setlength{\tabcolsep}{4pt}      
\renewcommand{\arraystretch}{1.15} 

\label{tab:train-hparams}
\begin{tabularx}{\linewidth}{>{\raggedright\arraybackslash}X c c c c c c c c}
\toprule
\textbf{Model} & \textbf{Optimizer} & \textbf{LR} & \textbf{$\beta$} & \textbf{$\varepsilon$} & \textbf{Warmup} & \textbf{MLM} &
\makecell{\textbf{Sliding}\\\textbf{Window (FA)}} &
\makecell{\textbf{ALiBi}\\\textbf{start size}} \\
\midrule
ModernBERT-base-PT & StableAdamW & 3e-4   & (0.90, 0.98) & 1e-06 & 6\% & 30\% & 256 & n/a \\
ModernBERT-base-VX & StableAdamW & 3e-4   & (0.90, 0.98) & 1e-06 & 6\% & 30\% & 256 & n/a \\
MosaicBERT-large   & StableAdamW & 2e-4 & (0.90, 0.98) & 1e-06 & 6\% & 30\% & n/a  & 1024 \\
\bottomrule
\end{tabularx}
\caption{Pretraining Hyperparameters.}
\end{table}

\textbf{Finetuning Hyperparameters}

\begin{table}[H]
\small
\setlength{\tabcolsep}{4pt}      
\renewcommand{\arraystretch}{1.15} 
\centering
\begin{tabular}{lccccccc}
\toprule
\textbf{Model} & \textbf{Optimizer}& \textbf{LR} & \textbf{beta} & \textbf{weight decay} & \textbf{epochs} & \textbf{seeds}\\
\midrule
ModernBERT-base-PT    &\texttt{adamw\_torch\_fused}&3e-5  &(0.9, 0.999)  &3e-6  & 20 & 4 \\
ModernBERT-base-VX    &\texttt{adamw\_torch\_fused}&3e-5  &(0.9, 0.999)   &3e-6  & 20  & 4   \\
MosaicBERT-large    &\texttt{adamw\_torch\_fused}&3e-5&(0.9, 0.999)  &3e-6 & 20  & 4  \\
PatentBERT          &\texttt{adamw\_torch}&2e-5  &(0.9, 0.999)  &1e-2  & 20 & 4 \\
ModernBERT-base    &\texttt{adamw\_torch}&5e-5  &(0.9, 0.999)   &5e-6  & 20  & 4   \\
\bottomrule
\end{tabular}
\caption{Finetuning Hyperparameters.}
\label{tab:hp-finetuning}
\end{table}

\section{Results}
\label{app:fullresults}
\begin{figure*}[h]
    \centering
    \includegraphics[width=\textwidth]{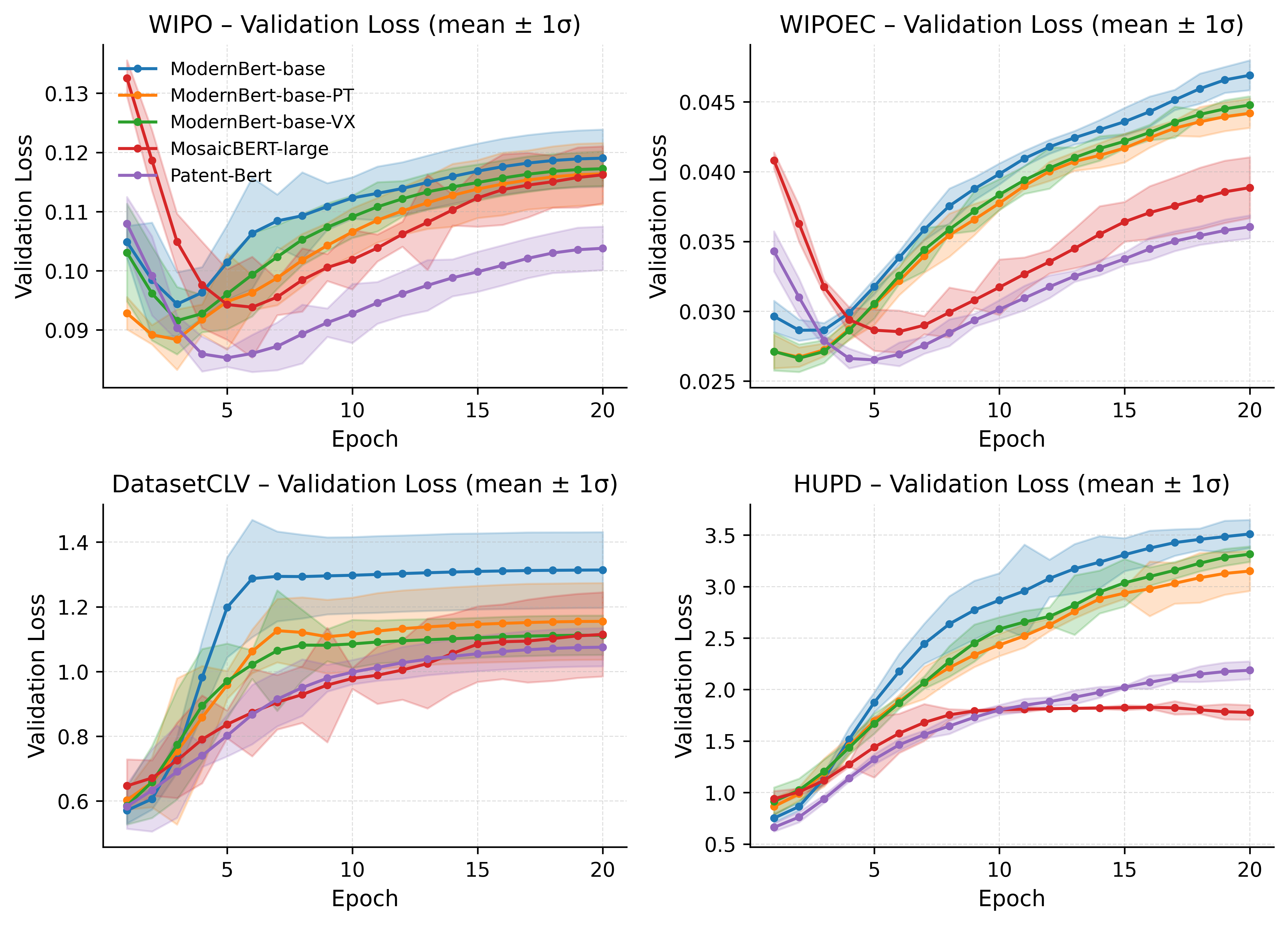}
    \caption{Validation loss curves over 20 epochs for four independent fine-tuning runs.}
    \label{fig:example-wide}
\end{figure*}

\begin{figure*}[t]
    \centering
    \includegraphics[width=\textwidth]{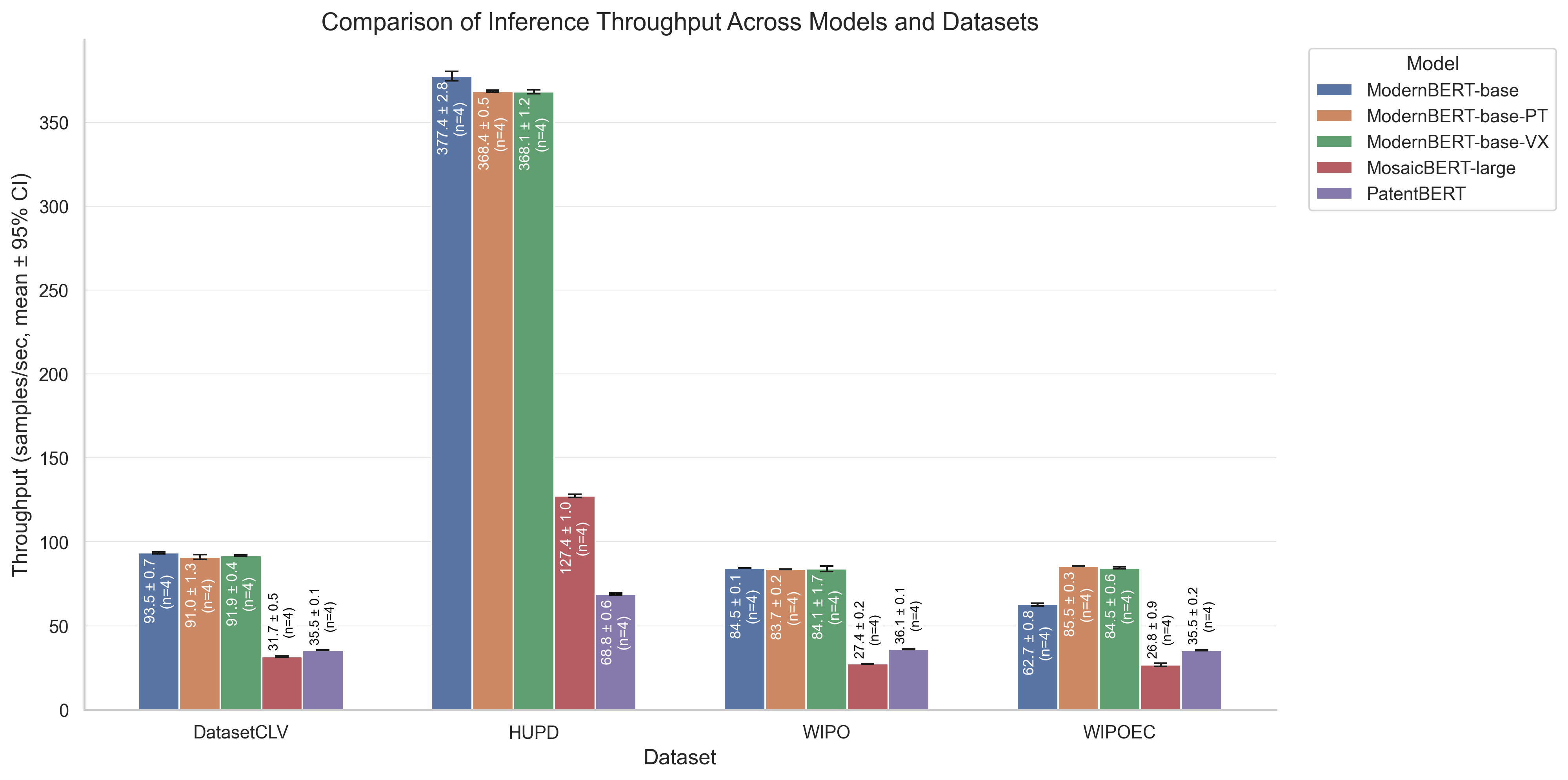}
    \caption[Test throughput across models and datasets]{\textbf{Test throughput} (\textit{samples per second}) across models and datasets, reported as mean \(\pm\) 95\% confidence interval computed over \textbf{4 random seeds} per (dataset, model). Confidence intervals are estimated using the \textit{Student's \(t\)} distribution. Higher values indicate greater inference efficiency.}
    \label{fig:all-datasets-models-throughput}
\end{figure*}

\begin{figure*}[t]
    \centering
    \includegraphics[width=\textwidth]{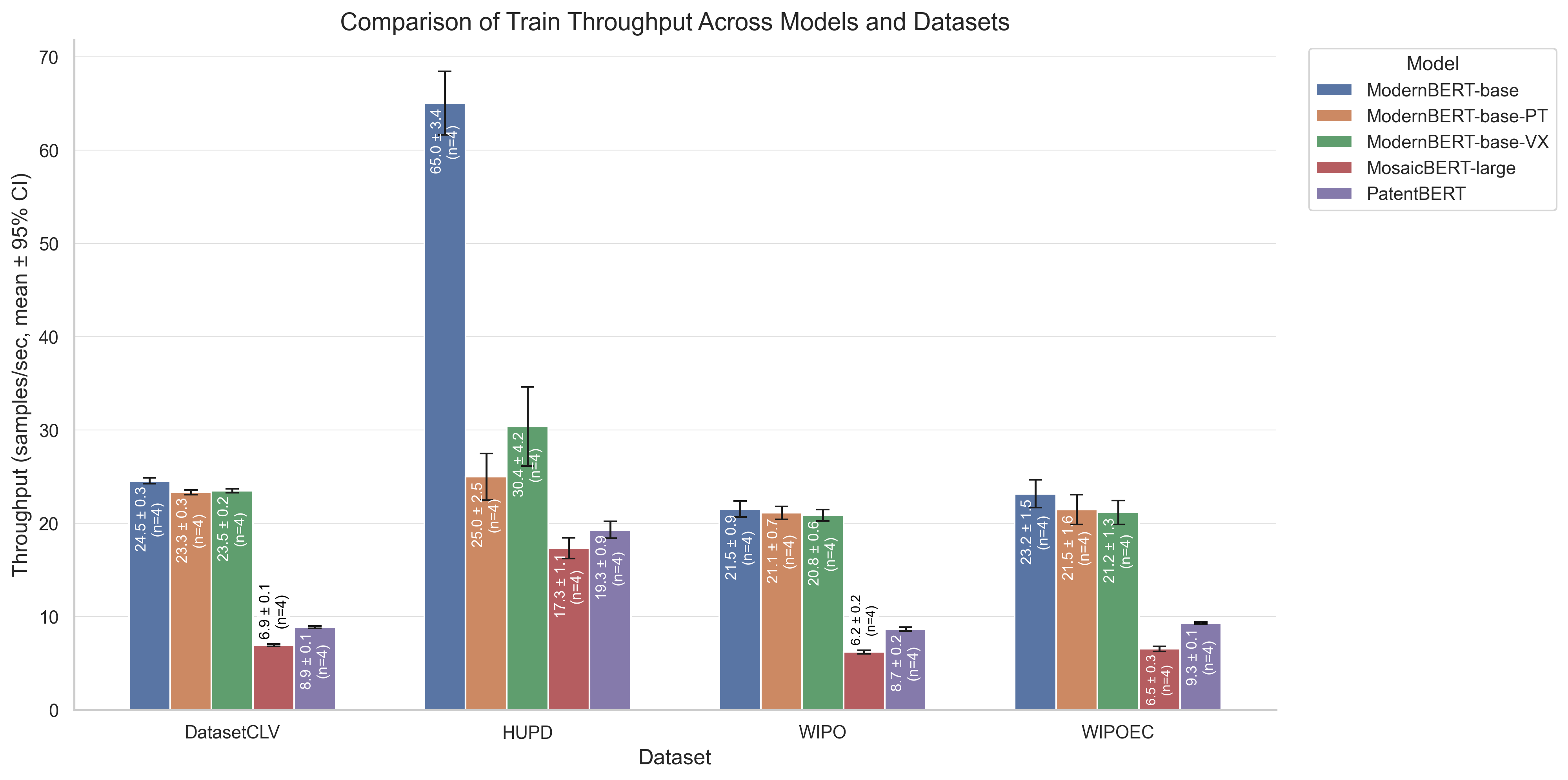}
    \caption[Train throughput across models and datasets]{\textbf{Train throughput} (\textit{samples per second}) across models and datasets, reported as mean \(\pm\) 95\% confidence interval computed over \textbf{4 random seeds} per (dataset, model). Confidence intervals are estimated using the \textit{Student's \(t\)} distribution. Higher values indicate greater inference efficiency.}
    \label{fig:all-datasets-models-training-throughput}
\end{figure*}

\end{document}